\documentclass[11pt]{article}
\usepackage{cmap}
\usepackage[T1]{fontenc}
\usepackage{lmodern}
\usepackage[utf8]{inputenc}
\usepackage[margin=1in]{geometry}
\usepackage{booktabs}
\usepackage{amsmath}
\usepackage{amssymb}
\usepackage{graphicx}
\usepackage{hyperref}
\usepackage{xcolor}
\usepackage{pgfplots}
\pgfplotsset{compat=1.18}
\usepackage{natbib}

\title{Playing $\log(N)$-Questions over Wikipedia Abstracts:\\
How Per-Round Errors Compound Under Information Asymmetry\\
\large A Technical Report}
\author{Peter Potash\\
\normalsize\href{mailto:pjpotash@gmail.com}{\texttt{pjpotash@gmail.com}}\\
\normalsize\url{https://github.com/ppotash/logn-questions}}
\date{\today}

\begin{document}
\maketitle

\begin{abstract}
We evaluate six frontier language models on the two-agent $\log(N)$-Questions
game of \citet{potash2019}. A questioner sees $N$ Wikipedia
lead paragraphs and must identify a secretly chosen target using exactly
$\log_2 N$ yes/no questions. An answerer sees only the target and the question,
and replies with one word. Both roles run on the same provider, so the game
measures how well a model communicates with itself across an information
asymmetry. We run 408 games over document sets of 4 to 1024 paragraphs at a
total API cost of \$363.

One model finishes well behind the others: Claude Opus 5 wins 28 of 68 games,
against 45 to 56 for GLM-5.3, GPT-5.6 Sol, Grok 4.6, Gemini 3.8 Flash and Kimi
K3. The leading five are only marginally separable. Pooling those five, win rate
declines with set size at $r=-0.973$ and is fit by a single per-round
reliability parameter. The form is $\text{win}=p^{\log_2 N}$ with $p=0.928$. It
presumes each round fails independently, which we test rather than assume: the
per-round failure rate is flat across the horizon ($\chi^{2}=10.5$,
$\mathrm{df}=9$, $p=0.31$), and a failure at one round does not raise the chance
of one at the next. Inverting the fit, coin-flip success needs $p \ge 0.933$ at
ten steps and $\ge 0.986$ at fifty, so the measured spread from 0.900 to 0.994
separates a 9-step horizon from a 115-step one.

We adjudicate each game's target and guess against three independent judges,
2{,}931 judgements each. Losses divide into answer errors and discrimination
failures in roughly equal measure overall, 47--55\% and 42--49\% depending on
the judge, with 2--4\% prediction errors. That aggregate describes no individual
model: Claude Opus 5's failures are $82\%$ answer errors ($4.6{:}1$,
$p=7\times10^{-5}$), while across the five leading arms discrimination leads by
a milder $1.73{:}1$ (63\%, 95\% CI $[53\%,73\%]$). Models almost never name a document their own evidence excludes. Neither
dominant failure is a breakdown of the channel between the agents: both are
single-agent mistakes, and what the two-agent structure supplies is the
conditions under which ordinary errors become undetectable, unrecoverable and
compounding. Both failure types are constant-rate processes: the per-round answer rate is flat across the horizon, and the
per-round probability that a question separates the target moves only from 0.974
to 0.968 between $N{=}16$ and $N{=}1024$, so failures accumulate because more
rounds must go right rather than because rounds get harder.
Measured per-round agreement, 0.900 to 0.994, reconciles with the fitted $p$,
the residual being the discrimination failures. Inter-judge agreement on
individual judgements is 97.8--98.7\%. Every unanimous answer error from the
weakest model was inspected: 32 of 34 are ``No'' answers, on properties stated
in the document's first sentence, under an instruction that explicitly warns
against defaulting to ``No''.

Information per question, estimated from answer balance, correlates with win
rate at $r=+0.88$. The only two models to extract a full bit per question are
the only two that partition on document titles, a strategy absent below
$N{=}32$ and used in a quarter of questions above it. Reasoning-token
expenditure varies $4.5\times$ across models with little relation to success,
and the trace grows as the candidate set shrinks without a matching gain in
reliability. Cost is uncorrelated with performance at $r=-0.05$: the two
cheapest arms win at \$0.36 and \$0.39 each, against \$2.46 for the most
expensive.

Two China-hosted providers refused documents in our corpus. Moonshot rejected
three of the 1024 paragraphs, on Taiwan's navy, a purged Chinese intellectual
and a detained Hong Kong activist; Z.ai rejected the third of these. We located them by bisection, iteratively
halving the document block and re-submitting until the triggering paragraph was
isolated, and report the refusals as a practical limit on assembled long
contexts. Our per-model ordering inverts published general-intelligence
leaderboards, on which the model finishing last here ranks at or near the top.
\S\ref{sec:external} sets out where independent benchmarks agree with our
result and where they do not.
\end{abstract}

\section{Introduction}

\citet{potash2019} propose a two-agent game in which a questioner distinguishes
among $N$ sentences by asking $\log_2 N$ yes/no questions of an answerer who
sees only the target. Both agents were trained end-to-end over a
Gumbel-softmax discrete channel and evaluated on small sentence sets.

We port the game to pretrained frontier models. This changes what it measures.
The original studied whether a communication protocol could be learned. Ours
studies whether one already exists: our agents never co-adapt, and must
coordinate zero-shot on a shared prior from pretraining. We run the game over
document sets from 4 to 1024 Wikipedia lead paragraphs.

The task makes three capacities jointly necessary, and produces one observable
when any of them fails. The questioner must partition a document set along a
semantic axis. It must predict how a second agent with strictly less context
will adjudicate that partition. And it must maintain a hypothesis state across
rounds without external scaffolding. A wrong final guess is consistent with
failure in any of the three, so decomposing the failure is the substantive
analytical problem.

Because both roles run on the same underlying model, failures cannot be
attributed to a capability gap between agents. What the arrangement supplies is
an information asymmetry with no feedback channel. The questioner never sees the
target, so a mistaken answer is indistinguishable from a correct one and
silently eliminates the target while the questioner continues to reason over a
set that no longer contains it. \S\ref{sec:comm} argues that the failures we can
categorise are single-agent mistakes, and that the asymmetry is what makes a low
per-call error rate expensive.

\paragraph{Relation to existing evaluation.} The task sits at the intersection
of three literatures. Long-context evaluation is dominated by retrieval-style
probes \citep{kamradt2023,liu2024lost}. RULER \citep{hsieh2024} broadened this
to include aggregation and multi-hop tracing, and SummHay \citep{laban2024}
argued that needle-style tests no longer separate frontier models. Our task is
an aggregation probe in that lineage, with one property those benchmarks lack:
the aggregate must be near-exact, because a misestimated partition is an
unrecoverable loss of information under a tight budget (\S\ref{sec:bits}).
Emergent-communication work has long studied paired agents solving reference
games \citep{lewis1969,foerster2016,lazaridou2020,kottur2017}; we adopt that
framing for two instances of a single pretrained model, not for agents trained
to co-adapt. Multi-hop question answering \citep{welbl2018,yang2018hotpotqa}
tests two to three dependent inferences. This task requires up to ten.

The main contributions of this work are as follows.

\begin{itemize}
\item A frozen, reproducible protocol for the game over document sets from
$N=4$ to $N=1024$, with nested document sets and prefix-extensible target
assignments.
\item Six complete model arms, 408 games, on identical documents and targets,
which makes every cross-model comparison paired.
\item An empirical fit of win rate to $p^{\log_2 N}$ with a single per-round
reliability parameter $p=0.928$ across nine set sizes.
\item An error decomposition requiring $O(1)$ rather than $O(N)$ adjudications
per round, showing that the two dominant failure types are constant-rate
processes and that the aggregate split masks two distinct failure profiles, validated across three judges, with a measured
self-preference discount of 40 to 46\% and inter-judge agreement of
97.8--98.7\%.
\item The finding that failures are single-agent mistakes rather than
breakdowns of a shared protocol, with one exception we can attribute to the
channel: questions encoding context the answerer cannot decode.
\item A judge-free measure of partition quality that correlates with win rate at
$r=+0.88$, and the observation that two models' opening questions are
systematically narrow at every scale we test.
\item A documented account of cross-provider engineering hazards, including
content filtering that made three encyclopedia paragraphs unusable and required
bisection to locate.
\end{itemize}

\section{Task}
\label{sec:task}

Let $D = \{d_1,\dots,d_N\}$ be an ordered document set and $t \in D$ the target.
The questioner observes $D$ in full; over $R = \log_2 N$ rounds it emits a
yes/no question $q_r$, and the answerer, observing only $(t, q_r)$, returns
$a_r \in \{\text{Yes},\text{No}\}$. After round $R$ the questioner names an
index and wins iff it is $t$'s.

\paragraph{No externalised hypothesis state.} The questioner receives the full
document set and the question/answer history each round, but never a running
list of viable candidates; it must re-derive the surviving set each round. This
follows the original architecture, where the hypothesis lived in a hidden state
vector.

\paragraph{The answerer never sees document indices.} It receives the target's
title and text but not its position, without which the game collapses into
integer bisection. Retaining the title leaves two non-semantic strategies
available: \emph{enumeration} (``is your document one of: A, B, C\dots?'') and
\emph{lexical bisection} (``does the title begin with a letter from A through
M?''). Both are decidable from a single document, which is the only constraint
the rules impose, and both permit exact partitions where semantic categories
cannot.

\paragraph{The prompt names no strategy.} The questioner is told the rules of
the game, what the answerer can see, and to aim for questions that divide the
remaining candidates evenly. It is not told that partitioning on titles is
permitted, nor that it is forbidden, nor that semantic questions are expected.
Neither enumeration nor lexical bisection is mentioned anywhere in either role's
prompt. The full text of both prompts is released with the code.

This is deliberate. Constraining the question form would measure compliance with
our constraint; suggesting a strategy would measure whether models follow a
hint. Leaving the space open means the strategy a model adopts is its own, and
differences between models are differences in what each treats as a natural way
to split a set. \S\ref{sec:aggregation} reports what they chose: three models
never partition on titles in 320 questions, two do so in roughly a quarter, and
the strategy is absent below $N{=}32$ across all six.

\section{Why this is a long-context and multi-step test}
\label{sec:longctx}

Long-context evaluation is dominated by retrieval-style probes --- needle in a
haystack \citep{kamradt2023}, multi-document QA \citep{liu2024lost} --- in which
the model must locate a small number of relevant spans within a large irrelevant
remainder. Such tasks are structurally well served by \emph{selective} sparse attention:
if only a few blocks are relevant, attending to the right few suffices. A
substantial line of work makes long context tractable precisely this way, from
fixed patterns \citep{child2019,beltagy2020,zaheer2020} to learned top-$k$
selection trained end-to-end \citep{yuan2025nsa,lu2025moba}.

$\log(N)$-Questions has the opposite structure. To ask a well-balanced opening
question over 1024 documents, the questioner must estimate what \emph{fraction}
of the whole set carries a candidate property. There is no small relevant
subset. Every document affects the answer equally, since each shifts the
partition by $1/N$. A model that attends to only part of its context cannot
know whether ``is the subject a person?'' splits the set 512/512 or 700/324,
and any such misestimate is an irrecoverable information loss under the tight
budget of \S\ref{sec:bits}.

Whether that is a real obstacle depends on the mechanism, and two that are
often grouped together behave differently here.
Top-$k$ selective attention gives each query a chosen subset of the context, so
a global proportion must be reassembled from partial views. Linear or recurrent
state maintains a fixed-size summary that has lossily absorbed every token.
The second has a different failure profile: it is weak on precise recall, not
on aggregate statistics. \citet{nawrot2025sparse}
provide the most direct evidence for the distinction, finding that selective
sparse attention degrades specifically on aggregation and high-scope tasks
while remaining safe for retrieval, and warning that ``sparsity levels safe for
retrieval tasks can cause failures on aggregation or multi-hop reasoning''.
Their study covers training-free sparsity on dense models, not trained
selection, and does not benchmark linear architectures --- so the clean
dichotomy is suggested rather than established. We note also that NSA and MoBA
both build in explicit global-recovery mechanisms, so citing them as evidence
of an inherent aggregation deficit would overstate the case.

It is worth being explicit about what each family predicts for the opening
question, since the predictions differ in \emph{sign} and our data can
distinguish them. Dense attention recovers the proportion in one pass, so
balance is limited only by whether the model computes the statistic.
Fixed-pattern sparsity gives each query a subset fixed in advance with a random
component, so a proportion estimated from the attended blocks should behave like
a sample proportion: correct on average, noisy in any instance. The prediction
is scatter around 50\%, not displacement from it. Learned top-$k$ selection
inverts this: selection is query-conditioned, so a query about property $X$
preferentially retrieves blocks matching $X$, the retrieved sample is enriched,
and the estimated fraction should be biased \emph{upward}. Linear or recurrent
state carries aggregate statistics cheaply, predicting good round-1 balance
paired with weak late-round recovery of specific survivors.

Our measurements fit none of these. The observed deficit is \emph{downward} for
all six models, with round-1 yes rates of 49, 49, 44, 41, 34 and 22\%, where
top-$k$ predicts upward bias and fixed-pattern predicts unbiased scatter. It is
also scale-invariant, as large at $N{=}8$ — where the whole set fits inside any
window any of these mechanisms would use — as at $N{=}1024$
(\S\ref{sec:aggregation}). A mechanism that engages only at long context cannot
explain a deficit already complete at a thousand-token prompt. The parsimonious
reading is that we are measuring a prior over what makes a good question, not a
limit on access to context: the models can see the whole set and choose to
partition it unevenly. That places the deficit with how a model
chooses to partition a set, not with its access to one.

Two further cautions. The architectures of the four proprietary models we
evaluate are undisclosed, so no observed behaviour can be attributed to a
specific attention mechanism; only Kimi K3 \citep{kimi2026k3} and GLM-5
\citep{zhipu2026glm5} have published technical reports. And the experiment that
would settle this needs open weights: run the same round-1 probe on one model
served under dense attention and under a sparse variant at matched context, and
compare the sign of the balance error. We present this section as motivation for
the task design, not as a hypothesis the results confirm.

The opening question is only half the task. Rounds beyond the first add a
second demand. The viable set is never given to
the questioner (\S2); it must be re-derived each round by re-applying every
prior question to all $N$ documents. At $N=1024$ the tenth round requires
holding two surviving candidates identified by re-adjudicating nine predicates
across a 126{,}000-token prompt in which they are not marked. This is
multi-step reasoning of a kind rarely tested: ten sequential, mutually
dependent inferences with no externalised scratchpad, where an error at any
step is silent and unrecoverable.

The task therefore measures aggregation and re-derivation over long context
rather than retrieval from it. RULER \citep{hsieh2024} includes aggregation
subtasks and is the closest precedent; the difference here is that the
aggregate must be near-exact and directly determines an action, so error is
compounded across rounds rather than scored once.

Whether this task can separate long-context aggregation from
question-generation ability more generally is left open.

\section{Corpus and protocol}

\subsection{Documents}

We reservoir-sample 4{,}000 lead paragraphs from the English Wikipedia
\texttt{wikimedia/wikipedia} 20231101 snapshot in a single streaming pass over
eight randomly chosen parquet shards. Acceptance requires the article's
\emph{first} paragraph block to pass a filter chain: 60--140 words, at least 2.0
sentence terminators per 100 words, no disambiguation or list-page markers, no
stripped-template artefacts, limited parenthetical clutter. The first-block
restriction matters: an earlier version scanned forward for the first
substantial block and silently fell into body sections whenever the lead was
short, producing documents that open with a section header glued to the text and
never state what the subject is. Roughly 15\% of a pilot sample was degraded
this way.

The 1024 documents used average 123 tokens (p50 118, p95 179, max 242), giving
an $N{=}1024$ document block of $\approx$126{,}000 tokens --- below the
200{,}000-token threshold at which several providers double input rates.

Because the questioner re-reads the full set each round, a single $N{=}1024$
game sends that block eleven times (ten questions plus a guess), for
$\approx$1.39M billable input tokens per game; the answerer's ten calls, each
one document and one question, contribute $\approx$3{,}500. Context
\emph{occupancy} is therefore constant at 126{,}000 tokens across every round
while the information that must be extracted from it changes completely: round
1 requires a global proportion over the whole set, round 10 requires locating
two surviving candidates within the same unmarked block. This is why prompt
caching (\S\ref{sec:eng}) dominates the cost structure, and why cached-token
counts of 1.3--2.0M per game appear in our logs.

\subsection{Nested sets and prefix-extensible targets}

Document sets are nested. Writing $D_N$ for the document set used at size $N$,
and $T_{1:k}$ for the first $k$ targets assigned to it, $D_{512} \subset
D_{1024}$ and so on down to $N=4$. Targets are emitted in bit-reversed order
over 64 buckets, so any power-of-two prefix is evenly stratified across the set
and $T_{1:8} \subset T_{1:16}$. Raising the run count therefore re-uses every
completed game. Each size draws from an independent
RNG. No target is reused, so $N=4$ contributes 4 games and every larger size 8,
for 68 per model.

All models see identical documents in identical order with identical targets,
making every cross-model comparison paired.

\subsection{Prompts and the communication channel}
\label{sec:prompts}

Both role prompts are provider-agnostic, built by a single module; providers
differ only in transport. Prompts split into system/cacheable/tail with the
document block in the cacheable segment, byte-identical across every round.

Prompts were revised four times during a pilot phase and then frozen before any
reported arm. None altered what either agent was asked to
reason about. Two corrected defects in our own instructions, and two altered
what the agents could assume about one another:

\begin{enumerate}
\item \textbf{World-knowledge licence.} Our first answerer instruction told it
to answer ``on the basis of the document alone''. This was a defect in the
prompt, not a model failure. A document describing a Spanish media group does
not contain the sentence ``Spain is in Europe'', so answering No to a Europe
question is the correct output of the instruction as written. Answer balance
under this wording was 21\% Yes across 24 calls where $\approx$50\% was
expected ($p=0.003$). The frozen prompt instead opens with ``Use ordinary world
knowledge'', states that the document will not always contain the answer
outright, and gives the Spain/Europe inference as a worked example.
\item \textbf{Answerer reasoning.} Reasoning had been disabled for the answerer
as a cost saving. One reply read ``No\dots\ wait. The document concerns
biathlon, a winter sport. Yes'' --- a reflexive No corrected only when the model
had room to think. Reasoning was restored.
\item \textbf{Presupposition prohibition.} The questioner produced questions of
the form ``is it X rather than Y'' where $Y$ appeared nowhere in the target
document, existing only because a \emph{different} candidate had that property.
The answerer could not decode why $Y$ was mentioned and answered No.
\item \textbf{Generalising the prohibition.} Naming only ``rather than'' was
insufficient; the next failure used ``(as opposed to equipment or machinery)''
on a document titled \emph{Task-oriented and relationship-oriented leadership}.
The rule was broadened to every phrasing of a contrast, and the answerer
instruction changed from adjudicating the contrast to ignoring it. Post-fix,
0 of 12 sampled questions were contrastive.
\end{enumerate}

The questioner is additionally told the answerer's rules verbatim, on the
grounds that any divergence between what the questioner assumes and what the
answerer is instructed to do is a direct tax on agreement.

\subsection{Reasoning protocol}
\label{sec:reasoning}

Each provider runs at \emph{its own} maximum-effort setting rather than a
matched token budget; matching counts would place models at non-native settings
and amount to tuning a knob to a metric. Token expenditure is therefore a result
rather than a controlled variable. The questioner runs at high effort, the
answerer at medium.

One documented exception: \textbf{Gemini 3.8 Flash runs at MEDIUM}. At HIGH it
spends 30{,}000--61{,}000 tokens reasoning on a single call, exhausting its own
64{,}000-token output budget before the visible answer completes --- on
truncated calls, visible output capped at 1{,}280 tokens while thinking consumed
the remainder. Raising the cap does not help, because thinking scales to fill
it. This is a property of the model's output budget on this task rather than a
configuration preference. Notably, Gemini at MEDIUM outperforms both other
models at HIGH (\S\ref{sec:results}), which is itself evidence against a
more-reasoning-is-better account.

\section{Cross-provider engineering}
\label{sec:eng}

A substantial fraction of the work was reconciling provider APIs. We document
the hazards because several fail silently and one invalidated a complete arm.

\subsection{Reasoning configuration is not portable}

\begin{table}[h]
\centering\small
\begin{tabular}{llll}
\toprule
Provider & Parameter & Values & Notes \\
\midrule
Anthropic & \texttt{thinking.type}+\texttt{output\_config.effort}
  & adaptive; low--max & on by default \\
OpenAI & \texttt{reasoning\_effort} & minimal--high & \\
Google & \texttt{thinkingLevel} & LOW/MEDIUM/HIGH & enum; budget deprecated \\
xAI & \texttt{reasoning\_effort} & low--xhigh & \\
Z.ai & \texttt{thinking.type} & enabled/disabled & binary \\
Moonshot & --- & --- & no exposed control \\
\bottomrule
\end{tabular}
\end{table}

Two API specifications were revised during our evaluation period. Claude Opus 5 rejects
\texttt{thinking.type}\,=\,\texttt{enabled} in favour of an adaptive mode with a separate
effort field; token budgets are ignored. Gemini 3 deprecates the integer
\texttt{thinkingBudget} for a \texttt{thinkingLevel} enum, and --- critically ---
a request sending only the deprecated budget receives the model's \emph{default}
level rather than an error. Adapters are therefore self-healing: each sends its
best guess, drops or substitutes any parameter a 400 names, and records the fact
in a \texttt{dropped\_params} field written to every result file.

\subsection{Sampling temperature is unavailable on three of six}

Claude Opus 5, GPT-5.6 Sol, and Gemini 3.8 Flash all reject or ignore
\texttt{temperature}. These arms run at provider defaults and are not
deterministic: we observed an identical game flipping win to loss between
identical configurations.

\subsection{Token accounting differs}

OpenAI reports reasoning tokens \emph{inside} \texttt{completion\_tokens}; xAI
reports them alongside. Taken at face value, xAI costs are understated by
$\approx$90\% on this workload; the diagnostic is \texttt{reasoning\_tokens}
exceeding \texttt{completion\_tokens}, observed as 3{,}853 against 308. Our
accounting normalises all usage fields to be disjoint.

\subsection{Stop-reason vocabularies are not portable}
\label{sec:stopreason}

This issue invalidated an entire experimental arm. Truncated replies must be treated as parse failures, since a
reply cut off mid-sentence still satisfies a format regex --- the fragment ends
its line. Our guard compared against the literal string \texttt{"max\_tokens"},
Anthropic's spelling. The observed vocabulary is \texttt{stop},
\texttt{end\_turn}, \texttt{STOP} and \texttt{MAX\_TOKENS}, with
\texttt{length} on OpenAI-compatible endpoints. Gemini's
\texttt{MAX\_TOKENS} never matched; the guard never fired; and a loose-parse
fallback lifted rhetorical questions out of abandoned reasoning and played them
as moves. Recorded examples include the question ``How many of those are
there?'' and two truncated guesses parsed as the integers 30 and 959.

The corruption affected 96 calls across 33 of 68 games, concentrated at large
$N$: all eight $N{=}1024$ games were contaminated and all eight were losses. The
reported $0/8$ was an artefact; the true figure after re-running at MEDIUM is
$6/8$. The guard now matches case-insensitively against all known spellings and
warns on any unrecognised stop reason, on the principle that a silently
unhandled terminal state is exactly how this survived a full arm.

\subsection{Provider content filtering blocks ordinary encyclopedia text}
\label{sec:filter}

Two of the six providers refused to process parts of our corpus. Both are
China-hosted. The refusals arrive as an opaque HTTP 400 naming the whole prompt,
with no indication of which span is responsible --- so at $N{=}1024$, a single
unacceptable paragraph makes a 126{,}000-token request unusable and gives no
clue why.

Z.ai (GLM-5.3) rejected every $N{=}1024$ prompt with error code 1301,
``potentially unsafe or sensitive content''. Moonshot (Kimi K3) rejected
$N{=}512$ and $N{=}1024$ with ``the request was rejected because it was
considered high risk''. Smaller sizes passed in both cases, which is expected:
more documents means more chances to include a triggering span.

\paragraph{Locating the triggers.} Because rejected requests are not billed and
a passing request can be truncated to a few tokens, bisection is cheap. We
narrowed the Moonshot block from 1024 documents to individual triggers in 105
API calls, then repeated after removing each. Moonshot's filter is not fully
deterministic --- ten consecutive identical requests were blocked 10/10, but two
earlier requests with the same prompt passed, and one of eight $N{=}1024$ games
completed normally while the other seven were refused. We therefore used an
asymmetric decision rule: a block is trusted immediately, a pass only after
three consecutive passes. Without that rule a single spurious pass sends the
search down the wrong half.

\paragraph{What was blocked.} Three of the 1024 documents:

\begin{itemize}
\item \emph{Cheng Kung-class frigate} --- guided-missile frigates of the
Republic of China Navy, built in Kaohsiung, Taiwan. Identified on Moonshot.
\item \emph{Luo Longji} --- Chinese politician and intellectual, called
``China's number two rightist'', purged in the Anti-Rightist Campaign, an early
promoter of human rights in the PRC. Identified on Moonshot.
\item \emph{Simon Cheng} --- Hong Kong activist detained by Chinese authorities
in 2019. Identified on both Moonshot and Z.ai.
\end{itemize}

Nothing else in 1024 randomly sampled Wikipedia lead paragraphs triggered
either filter. The three subjects are Taiwan's military, a purged Chinese
intellectual, and a Hong Kong dissident.

We say ``identified on'' rather than ``blocked by'' because the bisection
terminates once removing a document makes the set pass. Z.ai's search stopped
at Simon Cheng, so whether the other two documents would also trigger Z.ai was
never tested. The overlap on one document is established; the divergence on the
other two is not.

\paragraph{A control we ran, and its limits.} The Simon Cheng paragraph
contains the phrase ``soliciting prostitutes'', the charge Cheng denies, so a
sexual-content keyword is a competing explanation for the refusal. We tested
this on Z.ai by deleting the sentence containing that phrase, and the document
then passed.

The test is not decisive. The deleted sentence contained both a sexual-content
keyword (``soliciting prostitutes'') and a political narrative (``detained by
Chinese authorities \dots in West Kowloon station''). It therefore identified
the \emph{sentence} as the trigger but cannot determine which of the two
components caused the refusal. Separating them would require ablating each
independently. The other two documents contain no comparable confound.

\paragraph{Consequences for evaluation.} We substituted the blocked documents
with otherwise unused pool documents for the affected arms only, preserving $N$
and the round budget; this is recorded per game in the result files and noted in
\S\ref{sec:limits}. More generally: assembled long contexts drawn from a broad
corpus will, with probability growing in context length, contain something a
given provider will not process. The failure is opaque, is not mentioned in
context-window documentation, and --- in Moonshot's case --- is not reliably
reproducible. Any evaluation that assembles large contexts from open corpora
should expect this and should locate the cause rather than treating the
provider as unavailable.

\subsection{Prompt caching effectiveness varies by an order of magnitude}

Anthropic uses explicit \texttt{cache\_control} breakpoints; the others match
prefixes implicitly. On identical workloads with documents in a stable prefix,
cached tokens per $N{=}1024$ game were $\approx$2{,}008{,}000 (Anthropic),
1{,}310{,}000 (Gemini), and 127{,}000 (OpenAI), the last confirmed by OpenAI's
dashboard at an 8.6\% hit rate. Gemini's implicit caching does not engage below
$\approx$70{,}000 tokens but works well above it. This is an operational cost
difference independent of headline token rates.

\section{Results}
\label{sec:results}

\begin{table}[h]
\centering\small
\begin{tabular}{lrrrrrr}
\toprule
$N$ & Kimi K3 & Gemini 3.8 & Grok 4.6 & GPT-5.6 Sol & GLM-5.3 & Opus 5 \\
\midrule
4    & 4/4 & 4/4 & 4/4 & 4/4 & 3/4 & 4/4 \\
8    & 8/8 & 8/8 & 8/8 & 8/8 & 6/8 & 3/8 \\
16   & 8/8 & 8/8 & 7/8 & 7/8 & 6/8 & 1/8 \\
32   & 8/8 & 8/8 & 7/8 & 5/8 & 7/8 & 6/8 \\
64   & 8/8 & 6/8 & 5/8 & 5/8 & 6/8 & 2/8 \\
128  & 5/8 & 5/8 & 5/8 & 5/8 & 6/8 & 4/8 \\
256  & 6/8 & 5/8 & 5/8 & 6/8 & 3/8 & 3/8 \\
512  & 5/8 & 5/8 & 5/8 & 4/8 & 6/8 & 4/8 \\
1024 & 4/8 & 6/8 & 5/8 & 5/8 & 2/8 & 1/8 \\
\midrule
Total     & \textbf{56/68} & \textbf{55/68} & \textbf{51/68} & \textbf{49/68}
          & \textbf{45/68} & \textbf{28/68} \\
Win rate  & 82\% & 81\% & 75\% & 72\% & 66\% & 41\% \\
Cost      & \$68.72 & \$21.35 & \$70.02 & \$120.60 & \$16.26 & \$66.21 \\
\$/win    & 1.23 & \textbf{0.39} & 1.37 & 2.46 & \textbf{0.36} & 2.36 \\
Output tokens ($N{=}1024$) & 101{,}800 & 103{,}286 & 134{,}138 & 29{,}574 & 110{,}674 & 38{,}334 \\
Wall clock & 20.3\,h & 3.6\,h & 25.3\,h & 3.9\,h & 7.2\,h & 12.7\,h\,$^\dagger$ \\
$r$ vs $\log_2 N$ & $-0.88$ & $-0.83$ & $-0.90$ & $-0.82$ & $-0.63$ & $-0.46$ \\
\bottomrule
\end{tabular}
\begin{minipage}{\linewidth}\vspace{2pt}\footnotesize
$^\dagger$Claude Opus 5's wall clock is inflated by a provider outage
(\S\ref{sec:eng}) that forced long retry waits; its uninterrupted games ran
comparably to GLM-5.3's.
\end{minipage}
\caption{Six complete arms, 68 games each on identical document sets and
targets.}
\end{table}

The structure is a split, not a ranking. All five leading models beat Claude
Opus 5 (Fisher exact: $p=10^{-6}$ to $p=0.006$). Within the leading group only
the two extremes are marginally separable (Kimi $56/68$ vs GLM $45/68$,
$p=0.049$); every adjacent pair is indistinguishable. Reporting an ordering
inside that group would over-read eight games per cell.

Kimi K3 won every one of its 36 games at $N\le64$, the only model to play the
first five sizes without a loss.

Cost is uncorrelated with performance ($r=-0.05$). GLM-5.3 and Gemini 3.8 Flash
win at \$0.36 and \$0.39 against GPT's \$2.46, a six-fold efficiency difference
with no accuracy penalty. Wall clock varies more sharply still: Grok's arm took
25 hours against Gemini's 3.6, driven by reasoning volume rather than
throughput.

\begin{figure}[tbp]
\centering
\makebox[\textwidth][c]{%
\begin{tikzpicture}
\begin{axis}[
  width=11.5cm, height=7.5cm,
  xlabel={$\log_2 N$}, ylabel={win rate},
  xtick={2,3,4,5,6,7,8,9,10},
  xticklabels={4,8,16,32,64,128,256,512,1024},
  ymin=0, ymax=1.05, grid=major,
  legend style={font=\small, at={(1.02,1.0)}, anchor=north west,
                draw=black!30, cells={anchor=west}},
  legend cell align=left,
]
\addplot[thick,mark=pentagon*,brown] coordinates
  {(2,1.0)(3,1.0)(4,1.0)(5,1.0)(6,1.0)(7,0.625)(8,0.75)(9,0.625)(10,0.5)};
\addlegendentry{Kimi K3 (56/68)}
\addplot[thick,mark=triangle*,teal] coordinates
  {(2,1.0)(3,1.0)(4,1.0)(5,1.0)(6,0.75)(7,0.625)(8,0.625)(9,0.625)(10,0.75)};
\addlegendentry{Gemini 3.8 Flash (55/68)}
\addplot[thick,mark=diamond*,violet] coordinates
  {(2,1.0)(3,1.0)(4,0.875)(5,0.875)(6,0.625)(7,0.625)(8,0.625)(9,0.625)(10,0.625)};
\addlegendentry{Grok 4.6 (51/68)}
\addplot[thick,mark=square*,red] coordinates
  {(2,1.0)(3,1.0)(4,0.875)(5,0.625)(6,0.625)(7,0.625)(8,0.75)(9,0.5)(10,0.625)};
\addlegendentry{GPT-5.6 Sol (49/68)}
\addplot[thick,mark=o,orange] coordinates
  {(2,0.75)(3,0.75)(4,0.75)(5,0.875)(6,0.75)(7,0.75)(8,0.375)(9,0.75)(10,0.25)};
\addlegendentry{GLM-5.3 (45/68)}
\addplot[thick,mark=*,blue] coordinates
  {(2,1.0)(3,0.375)(4,0.125)(5,0.75)(6,0.25)(7,0.5)(8,0.375)(9,0.5)(10,0.125)};
\addlegendentry{Claude Opus 5 (28/68)}
\addplot[dashed,black,very thick,domain=2:10,samples=50] {0.928^x};
\addlegendentry{$p^{\log_2 N}$ fit, $p=0.928$}
\addplot[thick,dashdotted,black!70,domain=2:10,samples=60] {2^(-x)};
\addlegendentry{chance $=1/N$}
\end{axis}
\end{tikzpicture}}
\caption{Win rate against set size for all six models, with the fitted $p^{\log_2 N}$ curve ($p=0.928$) and the $1/N$ chance baseline. Pooled over the five leading models the fit gives $r=-0.973$.}
\label{fig:winrate}
\end{figure}
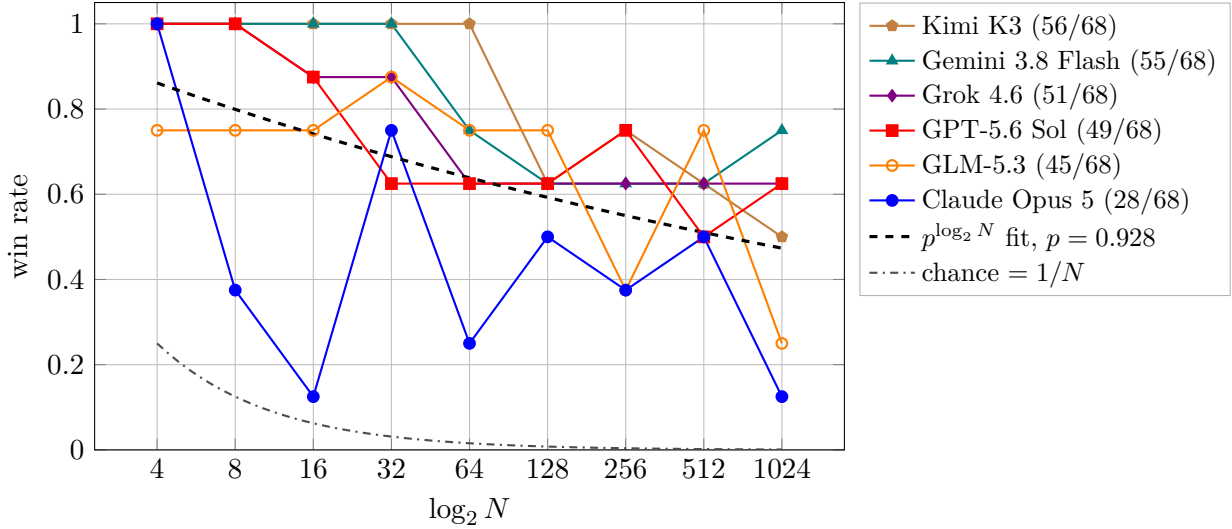

Because chance falls three orders of magnitude across this range, lift over
baseline rises to $768\times$ at $N=1024$ for the leading models.

\subsection{Win rate follows $p^{\log_2 N}$ with a single reliability parameter}
\label{sec:decay}

Pooling the five leading models --- Opus 5 is excluded here and treated separately
in \S\ref{sec:aggregation}, for reasons that section makes clear --- win rate
declines monotonically with set size:

\begin{center}\small
\begin{tabular}{lrrrrrrrrr}
\toprule
$N$ & 4 & 8 & 16 & 32 & 64 & 128 & 256 & 512 & 1024 \\
\midrule
Pooled win rate & 95\% & 95\% & 90\% & 88\% & 75\% & 65\% & 62\% & 62\% & 55\% \\
$p^{\log_2 N}$ fit & 101\% & 94\% & 87\% & 81\% & 75\% & 70\% & 65\% & 60\% & 56\% \\
Games & 19/20 & 38/40 & 36/40 & 35/40 & 30/40 & 26/40 & 25/40 & 25/40 & 22/40 \\
\bottomrule
\end{tabular}
\end{center}

The correlation with $\log_2 N$ is $r=-0.973$, and it is not an artefact of
pooling: every model's individual correlation is negative (Kimi $-0.88$, Gemini
$-0.83$, Grok $-0.90$, GPT $-0.82$, GLM $-0.63$, Opus 5 $-0.46$).

A compounding model fits the curve. Suppose each round succeeds independently
with probability $p$, meaning the questioner asks a usable question and the
answerer adjudicates it as intended. Then $\text{win} = p^{R}$ with
$R=\log_2 N$. Regressing $\log(\text{win rate})$ on $R$ gives $r=-0.973$ and an
implied $p=0.928$. A \emph{single} parameter reproduces performance across nine
set sizes spanning three orders of magnitude, and it is drawn as the dashed
curve above. The estimate is stable as arms are added: on five models it was
$p=0.931$, on six $p=0.928$.

This is the central quantitative result of the report. Questions asked at
$N{=}1024$ are not obviously harder than those asked at $N{=}32$. What grows
with the task is the number of opportunities to fail. At $p=0.93$ three rounds
are survivable and ten are not, and improving per-round reasoning quality
changes nothing unless it raises $p$. \S\ref{sec:comm} argues that the failures being
compounded are single-agent task mistakes amplified by the asymmetric channel,
not failures of mutual communication.

Two caveats. We do not test the independence assumption. A wrong answer early
may make later errors more or less likely, and these data cannot distinguish
the cases.
And $p=0.928$ is fit to five models pooled; individual values differ, and the
fit is descriptive rather than derived. The fitted curve also exceeds 1 at
$N{=}4$, which is a reminder that it is a regression rather than a model
derived from first principles.

\subsection{Information per question predicts the ordering}
\label{sec:bits}

The strongest single predictor we find is the empirical answer balance --- the
fraction of answerer replies that are ``Yes'' --- which is a direct estimate of
how much information each question extracts.

A yes/no question with response probability $p$ yields $H(p)$ bits. Identifying
one of $N$ documents requires $\log_2 N$ bits, and the questioner is allotted
exactly $\log_2 N$ questions, so the budget is tight by construction: any
$H(p) < 1$ is an unrecoverable deficit.

\begin{table}[h]
\centering\small
\begin{tabular}{lrrrrr}
\toprule
Model & Yes rate & $H(p)$ bits & bits at $N{=}1024$ & expected survivors & win rate \\
\midrule
Kimi K3          & 49\% & 1.000 & 10.00 & 1.00 & 82\% \\
GPT-5.6 Sol      & 49\% & 1.000 & 10.00 & 1.00 & 72\% \\
GLM-5.3          & 44\% & 0.990 &  9.90 & 1.07 & 66\% \\
Gemini 3.8 Flash & 41\% & 0.977 &  9.77 & 1.17 & 81\% \\
Grok 4.6         & 34\% & 0.925 &  9.25 & 1.70 & 75\% \\
Claude Opus 5    & 22\% & 0.760 &  7.60 & 5.23 & 41\% \\
\bottomrule
\end{tabular}
\caption{Round-1 answer balance, six completed arms. Survivors is $N \cdot 2^{-R\,H(p)}$, the expected
number of candidates remaining after a full game at this information rate. With
$R=\log_2 N$, a model extracting a full bit per question has
$2^{-R} = 1/N$ and expects exactly one survivor; anything less leaves more.}
\end{table}

A low answer balance has two possible causes that this statistic cannot
separate: a questioner asking narrow questions, or an answerer wrongly replying
``No''. \S\ref{sec:errors} separates them by adjudication and finds the latter
dominates for Opus 5: 32 of its 34 unanimous answer errors are ``No'' answers on
properties stated in the document. The figures below should therefore be read
as an upper bound on questioner-side deficit.

Opus 5 is short by 2.4 bits at $N=1024$, leaving $\approx$5 candidates on average
--- a one-in-five lottery at the final guess, by construction, however well it
tracks state. Opening questions selecting a fifth of the field rather than half
forfeit a quarter of each round's information, and the game provides no slack.
This is why Opus 5 is excluded from the compounding fit of \S\ref{sec:decay}: it
is not losing rounds at a rate $p$, it is starting each round with less than a
full bit available.

Across the six arms, $H(p)$ correlates with win rate at $r=+0.88$. The
relationship is not exact, with Grok 4.6 the exception, achieving 75\% on 0.925 bits,
better than GLM-5.3 manages on 0.990. We report it as a counterexample rather
than smoothing it. Partition quality is necessary but does not exhaust what
determines success. Kimi K3, which tops the table, also has the
best round-1 balance, tied with GPT --- and these are exactly the two models
that use lexical bisection (\S\ref{sec:aggregation}), a partition that can be
made exactly even by counting where a semantic category cannot.

Hand-adjudication corroborates the mechanism directly. On $N{=}16$ run 0, Opus 5
reduced $16 \to 9 \to 5 \to 3 \to 2$ with every answer correct --- splits of
0.56, 0.56, 0.60, 0.67 rather than 0.50 --- and lost the resulting coin flip.
The target never left the viable set; this is neither an answerer error nor a
state-tracking failure but an information deficit.

We emphasise that this metric is computed entirely from logged answers and
requires no judge model.

\subsection{A scale-invariant deficit in opening questions}
\label{sec:aggregation}

If a model's partitions are well calibrated, its answer balance should sit near
50\%. Deviation measures how badly it misjudged what fraction of the candidate
set carries the property it asked about. Splitting by round separates the
opening question --- which has no prior constraints to work from --- from later
rounds, which operate over a subset the model has already reasoned about:

\begin{table}[h]
\centering\small
\begin{tabular}{lrrrrr}
\toprule
Model & Round 1 & Rounds 2+ & $\Delta$ (R2+ $-$ R1) & $z$ vs 50\% (R1) & Win rate \\
\midrule
GPT-5.6 Sol      & 49\% & 46\% & $-3$ & $-0.16$ & 72\% \\
Kimi K3          & 49\% & 41\% & $-8$ & $-0.16$ & 82\% \\
GLM-5.3          & 44\% & 40\% & $-4$ & $-0.99$ & 66\% \\
Gemini 3.8 Flash & 41\% & 42\% & $+1$ & $-1.48$ & 81\% \\
Grok 4.6         & 34\% & 39\% & $+5$ & $-2.64$ & 75\% \\
Claude Opus 5    & \textbf{22\%} & 35\% & $\mathbf{+13}$ & $\mathbf{-4.62}$ & 41\% \\
\bottomrule
\end{tabular}
\caption{Yes rate by round; 50\% indicates a balanced partition. A positive $\Delta$ means the model is
better once the candidate set has been constrained. $z$ is against a balanced null over the 68
round-1 answers per model.}
\end{table}

Four of six models are flat or slightly worse once the candidate set is
constrained. They show no particular difficulty with the unconstrained opening
survey. Two models, Grok and Opus 5, are substantially worse on round 1 and
recover afterwards. Opus 5's opening questions select 22\% of the corpus rather than
half, a $4.6\sigma$ departure yielding $H(0.22)=0.76$ bits where a full bit is
available; Grok's 34\% is a $2.6\sigma$ departure. Under a balanced question,
one or fewer of eight targets falling on the ``yes'' side has probability
$0.035$; Opus 5 hit that on six of nine sizes.

\paragraph{Later rounds carry the residue of earlier errors.} The six models
split evenly on round-1 versus later balance, but part of the later-round figure
has a mechanical source. Once a wrong answer has eliminated the target, the
questioner partitions a set that no longer contains it, so a ``Yes'' requires
the target to happen to satisfy a predicate chosen for other documents.
Splitting rounds at the first adjudicated error shows this:

\begin{center}\small
\begin{tabular}{lrrr}
\toprule
Model & before (n) & after (n) & change \\
\midrule
Claude Opus 5    & 40.5\% (84) & 15.1\% (126) & $-25.4$ \\
GPT-5.6 Sol      & 53.8\% (13) & 31.8\% (44)  & $-22.0$ \\
Gemini 3.8 Flash & 37.5\% (16) & 19.2\% (26)  & $-18.3$ \\
Kimi K3          & 30.8\% (13) & 20.4\% (49)  & $-10.4$ \\
Grok 4.6         & 12.5\% (16) & 21.7\% (23)  & $+9.2$ \\
GLM-5.3          & 24.0\% (25) & 47.5\% (40)  & $+23.5$ \\
\midrule
Pooled           & 35.3\% (167) & 23.4\% (308) & $-12.0$ \\
\bottomrule
\end{tabular}
\end{center}

The pooled drop is $z=2.78$, $p=0.005$. Only the Opus 5 row is individually
significant, the per-model cells run from 13 to 126 rounds, and two models move
the other way, so the pooled figure is a property of the aggregate rather than
an effect established for each model. It does not account for the deficit
either: the pre-error rate is itself 14.7 points below balanced. Later-round
balance therefore mixes partitioning with how long the game has been dead, which
makes the opening the cleaner measure.

Round-1 balance is also the better predictor. Across the six arms, $H(p)$ from
round 1 correlates with win rate at $r=+0.88$, from later rounds at $r=+0.81$,
and the round-1/later gap itself at $r=-0.75$. Six points cannot
establish a functional form. Nor is balance sufficient: Gemini has a lower yes
rate than GPT and a higher win rate. Partition quality is necessary without
exhausting what determines success.

\paragraph{What the opening questions look like.} Statistics about answer
balance are hard to picture, so we show the actual round-1 questions. At
$N{=}16$ the models converge almost completely; at $N{=}1024$ they diverge
sharply.

\begin{table}[h]
\centering\footnotesize
\begin{tabular}{p{2.6cm}p{10.5cm}}
\toprule
\multicolumn{2}{l}{\textbf{$N=16$, run 1} --- all six ask the same question} \\
\midrule
Claude Opus 5 & Is the subject of the document an individual human being? \\
GPT-5.6 Sol & Is the subject a named individual being? \\
Grok 4.6 & Is the subject an individual human person? \\
GLM-5.3 & Is the subject of the document an individual person? \\
Kimi K3 & Is the subject an individual human or spirit? \\
Gemini 3.8 & Is the subject primarily associated with North America? \\
\midrule
\multicolumn{2}{l}{\textbf{$N=1024$, run 0} --- three distinct strategies} \\
\midrule
Claude Opus 5 & Is the subject an individual human being? \\
Grok 4.6 & Is the subject a person? \\
Gemini 3.8 & Is the subject located in, or does it originate from, the Americas? \\
Kimi K3 & Is the subject primarily associated with the United States? \\
GLM-5.3 & Is the subject a person or a work of creative media (such as a film,
  television program, album, song, book, or video game)? \\
GPT-5.6 Sol & \emph{Does the subject's English-language name begin with a
  numeral or a letter from A through G?} \\
\bottomrule
\end{tabular}
\caption{Round-1 questions. At $N{=}16$ five of six models ask a near-identical
question about personhood; the sixth asks about geography. At $N{=}1024$ the
same six diverge into semantic partitions of increasing compound complexity and,
for GPT, a lexical one.}
\end{table}

Two observations. The $N{=}16$ convergence is near-total. In run 1 all six
models open with essentially the same question, and in run 0 five of six do.
Whatever determines the opening move at small $N$ is a property of the corpus,
not of the model. Second, Claude Opus 5 asks the same opening question verbatim
at $N{=}16$ and at $N{=}1024$: ``Is the subject of the document an individual
human being?'' This is consistent with the scale-invariance established below,
and suggests it does not adapt its strategy to the size of the problem.

\paragraph{A strategy that emerges only above $N{=}32$.} Retaining document
titles leaves \emph{lexical bisection} available without the prompt mentioning
it (\S\ref{sec:task}): a question
like ``does the title begin with a letter from A through M?'' partitions
\emph{exactly}, at any set size, where semantic categories cannot. Counting
questions matching an alphabetic or ordinal pattern:

\begin{center}\small
\begin{tabular}{lrrrrrrrrr}
\toprule
$N$ & 4 & 8 & 16 & 32 & 64 & 128 & 256 & 512 & 1024 \\
\midrule
Lexical questions & 0\% & 0\% & 0\% & 0.8\% & 12.5\% & 9.8\% & 9.1\% & 7.4\% & 5.4\% \\
\bottomrule
\end{tabular}
\end{center}

The strategy is entirely absent below $N{=}32$ and appears abruptly at
$N{=}64$. Semantic partitions are evidently adequate for small sets --- a
natural category will divide 16 documents acceptably --- and stop being so
somewhere between 32 and 64, at which point some models switch to a partition
that is exact by construction.

Adoption is sharply divided. Over all questions at $N\ge64$:

\begin{center}\small
\begin{tabular}{lrr}
\toprule
Model & Lexical questions & Share \\
\midrule
GPT-5.6 Sol      & 83/320 & 25.9\% \\
Kimi K3          & 69/320 & 21.6\% \\
Grok 4.6         & 10/320 &  3.1\% \\
Claude Opus 5    &  0/320 &  0.0\% \\
Gemini 3.8 Flash &  0/320 &  0.0\% \\
GLM-5.3          &  0/320 &  0.0\% \\
\bottomrule
\end{tabular}
\end{center}

Three models never use it once in 320 questions. Two use it in roughly a
quarter. Those two, GPT-5.6 Sol and Kimi K3, are the same two whose round-1
answer balance is 49\%, giving $H(p)=1.000$. They are the only models in the
study to extract a full bit per question (\S\ref{sec:bits}). The mechanism is
straightforward: a partition on the first letter of a title can be made exactly
even by counting, where a semantic category cannot.

Lexical bisection is not necessary for good balance: GLM-5.3 reaches
$H(p)=0.990$ without using it. Nor is it sufficient for winning, since GPT ranks
fourth overall. The association between the strategy and the information-rate
measure is nonetheless exact across these six models, and it identifies one
mechanism by which a model can beat the granularity limit that semantic
categories impose on finite sets.

\paragraph{The deficit does not scale with $N$.} Opus 5's round-1 yes rate by
set size runs 75, 12, 12, 50, 12, 12, 12, 12, 25 percent from $N{=}4$ to
$N{=}1024$. The deficit is fully present at $N{=}8$, a prompt of roughly one
thousand tokens, and does not grow with the set. It is therefore not a
long-context effect, even though round 1 is the only round demanding an
unconstrained survey of the whole input (\S\ref{sec:longctx}).

\paragraph{Nor is it candidate-set size.} Comparing cells at matched candidate
count $N/2^{r-1}$, so that round 1 at $N{=}8$ sits beside round 4 at $N{=}64$,
Opus 5's round-1 balance is worse than its later-round balance by 19 points on
average, against $-1$ for GPT-5.6 Sol and $+6$ for Gemini 3.8 Flash. Round
position therefore matters independently of how many documents a partition must
balance. The comparison is confounded: the candidate count assumes even splits,
which is what Opus 5 fails to achieve, so its true later-round candidate sets
are larger than the matched cells imply. Resolving this needs the partition
adjudication described in \S\ref{sec:future}.

\paragraph{Summary.} Two models ask systematically narrow opening
questions --- covering roughly a fifth of the candidate set rather than half ---
at every scale tested, and this costs them a quarter of a bit per round in a
game that provides no slack. Neither context length nor candidate-set size
accounts for it, and no
sparse-attention mechanism we can construct predicts it either: top-$k$
retrieval should bias estimates toward the queried property, and fixed-pattern
sparsity should give unbiased scatter rather than a one-sided deficit
(\S\ref{sec:longctx}). The measurement itself is cheap, judge-free, and computable from
any run of this task, which is what we offer.

\subsection{Decomposing losses: answer, discrimination, and prediction}
\label{sec:errors}

Win/loss compresses three distinct failures into one bit. Separating them
requires knowing whether a given document satisfies a given question. It does
not require this for all $N$ documents. Two suffice: the target, and whatever
the model guessed. This reduces the adjudication burden from $O(N)$ per round
to $O(1)$, which is what makes the analysis affordable at $N{=}1024$. A total of 2{,}931
judgements covers every game in the study, at roughly \$2 per judge.

We define:

The three are mutually exclusive and each loss is attributed to the
\emph{first} thing that went wrong in it, so what follows are shares of games,
not counts of errors: a game with an answer error at round 2 is attributed there
whatever happens afterwards. Discrimination is labelled over a game's rounds
jointly rather than round by round, so per-type error totals are not recoverable
from these labels.

\begin{description}
\item[Answer error] An independent adjudication of (question, target)
disagrees with the answerer's reply. The target was eliminated by a wrong
answer, and every subsequent round was spent narrowing a set that could not
contain it. The round of first disagreement is the \emph{round of death}.
\item[Discrimination failure] Every answer was correct, but the guessed
document is \emph{also} consistent with all of them. The questions never
separated the two; the final guess was a lottery among survivors.
\item[Prediction error] Every answer was correct and the guessed document is
inconsistent with the evidence the model itself received. It had enough
information and chose wrong.
\end{description}

\paragraph{Three judges, and a self-agreement correction.} A judge that is also
one of the evaluated models will be easier on its own outputs. We therefore
adjudicated the full set three times, with Gemini 3.8 Flash, GPT-5.6 Sol, and
Grok 4.6, and report each model's error count under the judges that are not it.
We call this the \textbf{self-preference discount}, and it is measurable:

\begin{center}\small
$
\text{discount} \;=\; 1 - \dfrac{\text{errors under own judgement}}
{\text{mean errors under other judges}}
$
\end{center}

\noindent GPT scores 15 and 11 errors under the other two judges but 7 under
itself, a discount of $1 - 7/13 = 46\%$. Gemini scores 5 and 5 under others and
3 under itself, $1 - 3/5 = 40\%$. Two independent measurements of the same bias,
in the same range. Every per-model figure we report therefore excludes that
model's judgement of itself. The direction matches \citet{panickssery2024}, who
find a linear relationship between a model's ability to recognise its own
output and the strength of its self-preference.

Inter-judge agreement on individual judgements is high and consistent across
all three pairs: 97.8\%, 98.7\% and 97.8\% (Gemini--GPT, Gemini--Grok,
GPT--Grok), over roughly 2{,}900 shared pairs each. We do not read this as
evidence that the judgements are correct. \citet{panickssery2024} show that
frontier models recognise and favour their own generations, and the
LLM-as-judge literature more generally warns that agreement between two models
of the same generation may reflect shared systematic bias rather than access to
ground truth. High agreement establishes that the judgements are
\emph{reproducible}, not that they are right; a human-adjudicated subset would
be needed for the stronger claim, and we did not collect one.

These figures set a floor on resolution: measurable per-round reliability
above roughly $0.98$ cannot be distinguished from the judges' own disagreement
rate, which puts Grok's near-perfect score at the edge of what this method
resolves.

\begin{table}[h]
\centering\small
\begin{tabular}{lrrrrrr}
\toprule
Model & Gemini & GPT & Grok & self & mean (others) & empirical $p$ \\
\midrule
Claude Opus 5    & 32 & 32 & 34 & --- & 32.7 & 0.900 \\
GPT-5.6 Sol      & 15 &  7 & 11 &  7  & 13.0 & 0.967 \\
GLM-5.3          & 10 &  9 & 10 & --- &  9.7 & 0.976 \\
Kimi K3          &  7 &  8 &  8 & --- &  7.7 & 0.981 \\
Gemini 3.8 Flash &  3 &  5 &  5 &  3  &  5.0 & 0.988 \\
Grok 4.6         &  0 &  5 &  5 &  5  &  2.5 & 0.994 \\
\bottomrule
\end{tabular}
\caption{Games (of 68) containing at least one answer error, under each judge.
Each judge adjudicated the 124 losses; the GPT judge returned verdicts for 121,
the remaining three having failed with API errors.
Counts include games won despite an error, since those are agreement failures
even though they are not losses; 5 to 9 such games occur per judge. Opus 5 is
invariant across all three at 32--34, roughly $3.4\times$ the next model.
Empirical $p$ is the implied per-round agreement,
$(1-\text{err}/68)^{1/\bar{R}}$ with $\bar{R}=6.24$. Grok's zero under Gemini is
an outlier, not self-agreement: Grok scored 5 under its own judgement.}
\end{table}

\paragraph{Some games are won despite an answer error.} Five to nine games per
judge contain an adjudicated answer error and are nonetheless won: the
questioner eliminated the target on the record and named it anyway. These are
agreement failures but not losses, and we count them separately. Binning them
with losses would inflate the
loss total by four percent and misattribute wins.

\paragraph{Losses are split between two causes; the third barely exists.}
Aggregated over all models and sizes, and stable across judges:

\begin{center}\small
\begin{tabular}{lrrrr}
\toprule
Judge & Losses & Answer error & Discrimination & Prediction \\
\midrule
Gemini 3.8 Flash & 124 & 50\% & 48\% & 2\% \\
GPT-5.6 Sol      & 121 & 47\% & 49\% & 4\% \\
Grok 4.6         & 124 & 55\% & 42\% & 3\% \\
\midrule
Mean             & 123 & 51\% & 46\% & 3\% \\
\bottomrule
\end{tabular}
\end{center}

Prediction errors are negligible. Across roughly 123 losses under each judge,
only three to five involved a model naming a document its own evidence
excluded. Little goes wrong at the final inference step. The failures occur
upstream: either the answerer contradicts the question's intended partition, or
the questions never distinguish the target from a competitor.

\paragraph{The pooled split describes no individual model.} Claude Opus 5
supplies 40 of the 124 losses and roughly half of all answer errors. Separating
it from the rest gives two different pictures:

\begin{center}\small
\begin{tabular}{lrrrr}
\toprule
& Losses & Answer error & Discrimination & Prediction \\
\midrule
Five leading models & 84 & 30 (36\%) & 52 (62\%) & 2 \\
Claude Opus 5       & 40 & 32 (80\%) & \phantom{0}7 (18\%) & 1 \\
\bottomrule
\end{tabular}
\end{center}

The two groups are asymmetric in how strongly they separate. Opus 5's answer
errors outnumber its discrimination failures $4.6{:}1$, a share of 82\% among
the two ($p = 7\times10^{-5}$, two-sided binomial). Among the five leading
models the imbalance runs the other way but is much milder: discrimination
leads $1.73{:}1$, a 63\% share with a 95\% interval of $[53\%, 73\%]$
($p=0.020$). Discrimination is therefore the more common failure for those five
rather than the overwhelming one, and with roughly sixteen losses each the
per-model splits are too thin to report separately.

What the aggregate obscures is the contrast, not the magnitudes. A pooled
51/46 split implies a single population failing two ways in equal measure. The
data are one model failing predominantly at answering and five failing somewhat
more often at discriminating.

\paragraph{Both failure types are constant-rate processes.} Expressed per game,
both rise steeply with $N$: among the five leading models, discrimination
failures grow from 10\% of games at $N{=}16$ to 27.5\% at $N{=}1024$. The
relevant question is whether a \emph{round} gets worse, or whether there are
simply more of them.

For discrimination, write $q$ for the per-round probability that a question
separates the target from its remaining competitors. A game avoids
discrimination failure only by succeeding at every round, so $q = (1-d)^{1/R}$
with $d$ the per-game failure rate and $R=\log_2 N$:

\begin{center}\small
\begin{tabular}{lrrrrrrr}
\toprule
$N$ & 16 & 32 & 64 & 128 & 256 & 512 & 1024 \\
\midrule
Discrimination, per game & 10.0\% & 9.2\% & 11.7\% & 15.8\% & 24.2\% & 21.7\% & 27.5\% \\
$q$, per round           & 0.974 & 0.981 & 0.980 & 0.976 & 0.966 & 0.973 & 0.968 \\
\bottomrule
\end{tabular}
\end{center}

\noindent Per-game failure rises $2.8\times$ across this range while $q$ moves
by 0.006, from 0.974 to 0.968. A question separates about as well over 1024
documents as over 16. Discrimination failures accumulate because ten rounds must
go right instead of four, not because the questions degrade. The correlation
between $q$ and $\log_2 N$ is $r=-0.82$, but that is a correlation along a
nearly flat line; the per-game contrast between $N{=}16$ and $N{=}1024$ reaches
only $p=0.083$ (4/40 against 11/40).

Answer errors behave the same way, established directly rather than by
inference: the per-round rate is flat across the ten rounds of an $N{=}1024$
game (\S\ref{sec:decay}). Both halves of the decomposition are therefore
constant-rate processes compounding over a longer horizon, which is what
$p^{\log_2 N}$ requires.

\paragraph{What the errors of the outlying model look like.} We inspected all
34 of Claude Opus 5's unanimous errors under the frozen prompt.

\textbf{The errors are overwhelmingly ``No''.} Of 34 errors on which all three
judges agree the answerer was wrong, \textbf{32 are ``No'' answers} and two are
``Yes'' ($p = 3.5\times10^{-8}$ against a balanced null). This is a directional
bias, not a scatter of mistakes.

\textbf{They are not edge cases.} The property asked about is usually stated in
the document's first sentence:

\begin{center}\small
\begin{tabular}{p{5.2cm}p{7.2cm}}
\toprule
Question, answered ``No'' & Document opens \\
\midrule
Is the subject an individual human being? &
  ``Walter Lee Gaines (17 March 1881 -- 20 November 1950) was a pioneer of dairy
  science\dots'' \\
Is the subject a woman? &
  ``Daniella Alonso is an American actress and former fashion model.'' \\
Is the subject Spanish? &
  ``Vocento, S.A.\ \dots is a Spanish mass media group.'' \\
Is the subject a musician or singer? &
  ``Liu Yuning \dots is a Chinese singer, actor, and the lead singer of Modern
  Brothers.'' \\
Is the subject a sports team? &
  ``The United States women's cricket team is the team that represents the
  country\dots'' \\
Was the subject formerly an administrative municipality that has since been
  dissolved? &
  ``H\o v\aa g \dots is a former municipality \dots existed from 1865 until its
  dissolution.'' \\
\bottomrule
\end{tabular}
\end{center}

``Is the subject an individual human being?'' is answered ``No'' about a person
in seven separate games. No inference beyond the text is required in any of
these cases.

\textbf{Context-faithfulness does not account for them.} Models differ in how
strictly they ground answers in provided text
\citep{longpre2021,xie2024conflict}, and Anthropic's public posture sits far
toward the grounding end: a Citations API built so that answers are anchored in
source documents, hallucination guidance instructing users to restrict the model
to provided material, and a published constitution setting an elevated honesty
bar \citep{bai2022cai}. Such a posture predicts declining inferences \emph{one
step outside} the document. These errors are the opposite case: they contradict
the document's own first sentence, where no inference is required at all.

\textbf{What the data show.} One model's answerer carries a strong bias toward
``No'', which persists under an instruction that says verbatim: ``Yes and No are
equally acceptable answers: do not fall back on No when uncertain.'' The bias is
visible in three independent measurements: a 22\% round-1 yes rate
(\S\ref{sec:bits}), the lowest agreement rate of any model at $p=0.900$, and the
32:2 direction of its errors. We have no mechanism to offer. Generic accounts of
post-training damaging calibration \citep{kadavath2022} apply to all six models
and cannot explain a difference between them, and the architecture of this model
is undisclosed (\S\ref{sec:limits}).

The bias is not confined to the weakest model: round-1 yes rates are below 50\%
for all six (49, 49, 44, 41, 34, 22\%), so some pull toward ``No'' appears
general and differs across models by degree.

\paragraph{How the rate compares to published grounded-task benchmarks.} A
per-round answer-error rate of 0--3.5\% for the five leading models is low
against the closest published comparison. Vectara's hallucination leaderboard
\citep{vectara2026} scores whether a summary of a short document contains claims
unsupported by it, which shares this task's structure: one short document, one
output that must remain faithful to it. Earlier-generation models score well
there, with GPT-4o at 1.5\%, Llama-3.1-405B at 3.9\% and Claude 3.5 Sonnet at
4.6\% \citep{bang2025hallulens}. Current reasoning models score considerably
worse: GPT-5, Claude Sonnet 4.5, Grok-4 and Gemini-3-Pro all exceed 10\% on the
revised dataset, one Grok variant reaches 20.2\%, and Gemini-3-Pro is reported
at 13.6\% \citep{vectara2026}.

The comparison is not exact. Summarising a document requires many claims to
remain supported, while our answerer makes one binary judgement, so a lower rate
is expected. But the ordering is informative in both directions. Our five
leading models sit below published rates for comparable reasoning models, and
Opus 5's 3.3--20.8\% sits inside that published range rather than outside it.
Neither figure is anomalous. What the comparison establishes is that a task this
constrained still does not reach zero, and that the residual is enough to
dominate outcomes once ten rounds must all go right.

\paragraph{The two groups fail differently at every size.} Setting the per-round
figures side by side:

\begin{center}\small
\begin{tabular}{lrr}
\toprule
& Five leading models & Claude Opus 5 \\
\midrule
$q$ (discrimination), range        & 0.966--0.992 & 0.931--1.000 \\
$q$ against $\log_2 N$             & $r=-0.85$    & $r=+0.07$ \\
Answer error per judged round      & 0--3.5\%     & 3.3--20.8\% \\
Answer error per game at $N{=}16$  & \textbf{0\%} & \textbf{62.5\%} \\
\bottomrule
\end{tabular}
\end{center}

Opus 5's discriminating power does not degrade with set size at all, and its
failures are overwhelmingly answers rather than partitions. The $N{=}16$ row is the sharpest
statement of the difference available in this data: across 40 games, the five
leading models produce no answer errors whatever, while Opus 5 produces one in
five of eight games.

\paragraph{Total losses by size and cause.}

\begin{figure}[tbp]
\centering
\makebox[\textwidth][c]{%
\begin{tikzpicture}
\begin{axis}[
  width=11cm, height=6cm, ybar stacked, bar width=15pt,
  xlabel={$N$}, ylabel={losses (mean of 3 judges)},
  symbolic x coords={4,8,16,32,64,128,256,512,1024},
  xtick=data, ymin=0,
  legend style={font=\small, at={(1.02,1.0)}, anchor=north west,
                draw=black!30, cells={anchor=west}},
  legend cell align=left,
]
\addplot+[fill=blue!55,draw=blue!70] coordinates
  {(4,0.7)(8,6.0)(16,5.0)(32,2.7)(64,9.7)(128,8.7)(256,9.0)(512,8.7)(1024,12.0)};
\addlegendentry{answer error}
\addplot+[fill=orange!65,draw=orange!80] coordinates
  {(4,0.3)(8,1.0)(16,6.0)(32,4.3)(64,5.7)(128,7.7)(256,10.7)(512,9.3)(1024,11.7)};
\addlegendentry{discrimination failure}
\addplot+[fill=black!45,draw=black!60] coordinates
  {(4,0.0)(8,0.0)(16,0.0)(32,0.0)(64,0.7)(128,1.3)(256,0.3)(512,1.0)(1024,0.7)};
\addlegendentry{prediction error}
\end{axis}
\end{tikzpicture}}
\caption{Losses by set size and cause, averaged over three judges. Total losses grow with $N$ while the share attributable to each cause is stable.}
\label{fig:losses}
\end{figure}
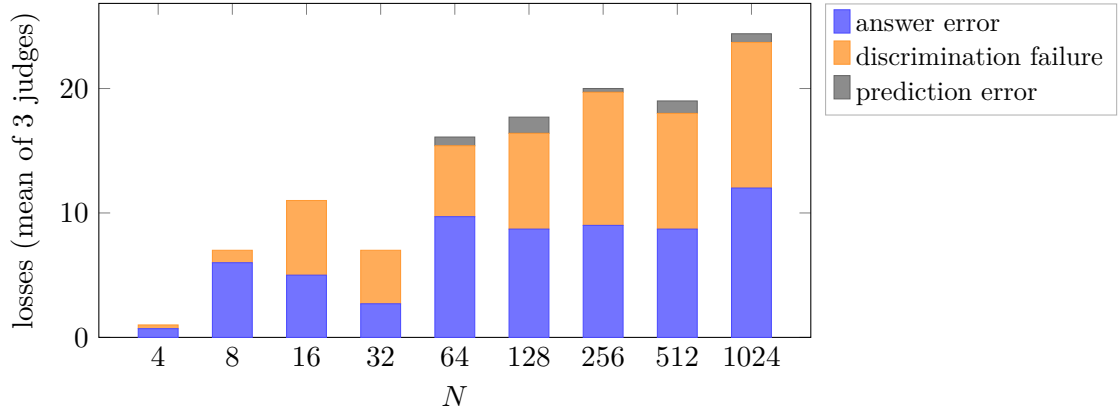

Total losses grow with $N$, as the win-rate curve requires, and the share
attributable to each cause is stable: pooling $N\le32$ against $N\ge256$ gives
answer shares of 54\% and 48\% (Fisher exact, $p=0.82$). Individual cells are
small, with seven losses in total at $N{=}8$. Prediction errors are absent
below $N{=}64$ and appear sporadically above it, one or two games per size.

\paragraph{Where the target dies.} Among games lost to answer error, expressed
as a fraction of the round budget: GPT 0.27--0.35, Kimi 0.34--0.52, Gemini
0.38--0.50, Grok 0.53, GLM 0.51--0.54, Opus 5 0.55--0.62. Opus 5 not only fails
more often but fails later, leaving fewer rounds' worth of useful work behind
--- though since the questioner cannot detect the failure, every round after the
death is wasted regardless of when it occurs.

\subsection{Signature collapse}

With $R$ questions, perfect play is a bijection: every document receives a
distinct $R$-bit answer signature. Two targets sharing a signature are
structurally indistinguishable and the final guess is a lottery.

Pooled across all completed arms, games whose signature was shared with another
game won 18/41 (44\%), against 121/173 (70\%) for games with a unique signature
($p=0.003$). At $N{=}16$ the contrast is stark: Opus 5 produced four distinct
signatures over eight targets --- with \texttt{FFFF} recurring four times and
winning none --- while Gemini produced eight and won all eight, and GPT produced
eight and won seven.

\begin{table}[h]
\centering\small
\begin{tabular}{llll}
\toprule
Model & $N{=}16$ signatures & Distinct & Won \\
\midrule
Claude Opus 5 & \texttt{FFFF}$\times$4, \texttt{FTFF}$\times$2,
  \texttt{FTTF}, \texttt{TTFF} & 4 & 1/8 \\
GPT-5.6 Sol & 8 distinct patterns & 8 & 7/8 \\
Gemini 3.8 Flash & 8 distinct patterns & 8 & 8/8 \\
\bottomrule
\end{tabular}
\end{table}

The metric compares signatures among the eight tested targets, not among all
$N$ documents. At $N=1024$ eight targets cover
1\% of the set, so distinctness is near-guaranteed by chance and the metric
loses power --- Opus 5 shows 8/8 unique at $N=1024$ yet wins only once. The
pooled effect above is therefore driven by $N \le 32$, and signature collapse is
best read as the small-$N$ consequence of the information deficit in
\S\ref{sec:bits} rather than as an independent explanation.

\subsection{Reasoning expenditure does not track performance}
\label{sec:trace}

At nominally equivalent settings, per-game output tokens at $N{=}8$ span
$11\times$: GPT 446, GLM 950, Opus 5 824, Kimi 2{,}644, Gemini 4{,}272, Grok
5{,}054. At $N{=}1024$ the spread is $4.5\times$: GPT 29{,}574, Opus 5 38{,}334,
Kimi 101{,}800, Gemini 103{,}286, GLM 110{,}674, Grok 134{,}138. Gemini, run at
MEDIUM for the reasons in \S\ref{sec:reasoning}, outperforms two models run at
HIGH. The data are inconsistent with a simple more-reasoning-is-better account
of the kind that motivates test-time compute scaling
\citep{wei2022cot,snell2024}, and consistent with \S\ref{sec:bits}: what matters
is the partition a question induces, not the deliberation behind it.

\paragraph{The trace grows as the candidate set shrinks.} Across the ten rounds
of an $N{=}1024$ game, mean questioner output rises from 5{,}713 tokens at round
1 to 11{,}413 at round 9 ($r=+0.69$), while the answerer moves only from 43 to
49. That direction is what a rational agent would produce. The candidate set
shrinks but the work of finding it grows: at round 9 the questioner re-applies
eight predicates across 1024 unmarked documents to recover two survivors, while
the answerer's job is one document and one question at every round.

The growth does not convert into outcomes, and the per-model profiles diverge
without sorting the ordering:

\begin{center}\small
\begin{tabular}{lrrl}
\toprule
Model & R1 & R9 & profile \\
\midrule
GLM-5.3          &  1.9k & 29.2k & ramps $15\times$ \\
Claude Opus 5    &  0.3k &  4.8k & ramps $18\times$ from a very low base \\
Kimi K3          &  5.9k &  8.5k & ramps $2.4\times$ \\
Grok 4.6         &  8.3k & 12.6k & flat and high \\
GPT-5.6 Sol      &  1.9k &  2.0k & flat and low \\
Gemini 3.8 Flash & 16.0k & 11.4k & starts high, falls \\
\bottomrule
\end{tabular}
\end{center}

The pooled $+434$ tokens per round is an average over qualitatively different
policies, not a shared law. For an agent budgeting its own trace, elapsed rounds
are a poor proxy for how much thinking helps.

\paragraph{Two observations we flag as suggestive.} Opus 5 spends 296 tokens on
the round-1 survey of 1024 documents against Gemini's 16{,}021, and has the
study's worst round-1 partition at $H(p)=0.760$. Across all six models the
correlation between round-1 log tokens and round-1 $H(p)$ is $r=+0.69$, but it
is driven entirely by that one point and reverses to $r=-0.49$ without it. With
six models this establishes nothing beyond the observation about Opus 5 itself.
Separately, GPT-5.6 Sol reaches $H(p)=1.000$ on the smallest traces in the
study, consistent with the lexical-bisection strategy of
\S\ref{sec:aggregation}: an exactly-even partition on first letters is cheap to
compute where a balanced semantic category is not.

\section{How single-agent errors compound under asymmetry}
\label{sec:comm}

\S\ref{sec:errors} attributes losses to two causes, and neither is a failure of
communication in the ordinary sense. A discrimination failure is the questioner
choosing questions that do not separate the target from a competitor; every
answer in such a game is correct, and the answerer plays no part in it. An
answer error is the answerer reading one short paragraph and judging it wrongly;
the questions admit clear answers, so no meaning is lost in transit. Both are
single-agent mistakes. So is the third, and rarest: a questioner naming a
document its own evidence excludes.

What the two-agent structure supplies is not a new kind of error but the
conditions under which ordinary errors become expensive.

\paragraph{Errors are undetectable.} The questioner never sees the target, so a
wrong answer is indistinguishable from a right one. There is no contradiction to
notice, no confidence signal to read, and no later evidence that conflicts:
subsequent answers are about a document that was already eliminated, and they
arrive as ordinary Yes/No replies. In a single-agent formulation of the same
task, a model that mistakenly judged an actress not to be a woman would have the
document in front of it and might catch the error on re-reading. Here it cannot.

\begin{figure}[tbp]
\centering\small
\fbox{\begin{minipage}{0.93\textwidth}
\textbf{A perfectly played game, lost to one wrong answer.} Claude Opus 5,
$N{=}32$, target [1] \emph{Grupo Vocento} (``a Spanish mass media group'').
Split sizes are the true partitions, adjudicated by hand against all 32
documents.

\medskip
\begin{tabular}{clrl}
\toprule
R & Question & Split & Answer \\
\midrule
1 & Is the subject associated with the United States?  & \textbf{16 / 16} & No \quad$\checkmark$ \\
2 & Is the subject associated with a country in Europe? & 8 / 8 & \textbf{No} \quad$\times$ \\
3 & Is the subject associated with Asia or Oceania?     & 4 / 3 & No \\
4 & Is the subject a species?                            & 2 / 2 & No \\
5 & Is the subject an individual human?                  & 1 / 1 & No \\
\bottomrule
\end{tabular}

\medskip
The questioner executed a textbook binary search: $32 \to 16 \to 8 \to 4 \to 2
\to 1$, with an exact 16/16 first split. Its final guess, \emph{Sea Shepherd
II}, is the \emph{uniquely correct} document given the five answers it
received. It lost because Vocento is Spanish, Spain is in Europe, and the
answerer, the same model holding only that one document, said No.

The questioner cannot detect this. There is no contradiction to notice: rounds
3, 4 and 5 were spent bisecting a set that could no longer contain the target,
and every question in them was well chosen.
\end{minipage}}
\caption{An \textbf{answer error}: the model incorrectly excludes the target at
round 2, and every later round partitions a set that cannot contain it.
Round-of-death: 2. This game was played during the pilot, under the first
answerer instruction, which forbade world knowledge, so the round-2 answer is
what that instruction required. It is shown to illustrate how one answer
propagates through the remaining rounds, not as evidence about the model.}
\label{fig:vocento}
\end{figure}

\paragraph{Errors are unrecoverable.} The game has exactly $\log_2 N$ questions
and no mechanism for revisiting an answer. Once an answer eliminates the target,
every remaining round partitions a set that cannot contain it
(Figure~\ref{fig:vocento}). The questions
asked in those rounds are often good ones, and \S\ref{sec:errors} shows the
target dies at 0.27 to 0.62 of the round budget depending on the model, so
roughly half the work in a lost game is expended after the game is already
decided.

\paragraph{Errors compound.} \S\ref{sec:decay} fits
$\text{win}=p^{\log_2 N}$ with $p=0.928$, and \S\ref{sec:errors} shows both
failure types are constant-rate processes: the per-round answer rate is flat
across the horizon, and the per-round separating power $q$ moves only from 0.974
to 0.968 between $N{=}16$ and $N{=}1024$. A per-call error rate of 2--3\%, which
is low by the standards of published grounded-task benchmarks
(\S\ref{sec:errors}), becomes a 25\% per-game failure rate at $N{=}1024$ purely
because ten rounds must all go right. The asymmetry does not make any individual
judgement harder. It removes every opportunity to catch one.

\paragraph{One failure is communicative.} The exception is a question that
encodes context the answerer cannot decode. ``Is the subject primarily
associated with Spain rather than Italy?'' mentions Italy only because a
different candidate is Italian; ``Is the subject related to human behavior,
management, or leadership (as opposed to equipment or machinery)?'' carries the
same intrusion, and was asked of a document titled \emph{Task-oriented and
relationship-oriented leadership}. Both were answered No. Neither is
unanswerable: a reader holding only the target document can determine that a
Spanish company is Spanish, and that an article about leadership concerns
leadership. But the questioner has leaked its own view of the candidate set into
an utterance for an interlocutor who cannot interpret it, which is a
theory-of-mind lapse in the sense studied by \citet{kosinski2024} and critiqued
by \citet{ullman2023}.

Prohibiting the construction in the questioner prompt eliminated it: 0 of 12
sampled questions were contrastive afterwards (\S\ref{sec:prompts}). This is the
one prompt revision that changed agent behaviour rather than correcting an error
in our own instructions, and the only failure in the study we can attribute to
the channel rather than to an agent.

\paragraph{Relation to emergent communication.} The setting is a Lewis
signalling game \citep{lewis1969} of the kind studied in the
emergent-communication literature \citep{foerster2016,lazaridou2020}, with one
difference that matters for interpreting the above: our agents do not co-adapt.
They arrive with a shared prior from pretraining rather than a learned protocol.
That the protocol largely works, and that failures are individual rather than
communicative, is itself the result. Where \citet{kottur2017} find that agents
trained to co-adapt develop opaque codes, two instances of a pretrained model
coordinate in natural language well enough that we could not identify a
systematic class of miscommunication beyond the contrastive questions above.

\paragraph{What we did not measure.} Our adjudication compares each answer to a
third-party judge, which tests whether the answer is \emph{correct}. It does not
test whether the answerer's judgement matches the \emph{questioner's intent},
which is the quantity a communication account would need. The questioner's
reasoning traces frequently state the intended partition, so this is computable
from the existing logs and would distinguish an answerer that is wrong from an
answerer that understood a different question. We flag it in
\S\ref{sec:future}.

\section{Relation to published benchmarks}
\label{sec:external}

Our ordering does not match published general-intelligence leaderboards, and
the discrepancy is worth stating plainly rather than leaving for a reader to
discover.

\paragraph{The outlier is a top-ranked model elsewhere.} At release, Claude Opus 5 took first place on the Artificial Analysis Intelligence Index
(61 at launch, rescored to 63), holding the lead until Claude Fable 5.1 scored
66. Our leading models sit \emph{below} it on that index: Kimi K3 at 57.1,
Gemini 3.8 Flash at 59, Grok 4.6 at 61. On aggregate intelligence, our result
is inverted relative to published rankings.

Two things follow. First, this should be read as a \emph{task-specific}
inversion, not a general capability ranking: we measure one narrow ability
under one protocol with eight games per cell. Second, Opus 5 is Anthropic's
mid-priced flagship rather than its top model --- the Fable and Mythos tiers sit
above it --- so ``the frontier Anthropic model'' is not what we tested, and
comparisons that place Opus 5 against other vendors' top tiers are not
like-for-like. Our protocol of one flagship per provider is defensible but
imperfect for exactly this reason (\S\ref{sec:limits}).

\paragraph{Where the published evidence agrees.} On long-context
\emph{comprehension} rather than retrieval, the ordering matches. Kimi K3 leads
the Artificial Analysis Long Context Reasoning benchmark at 88.7\%, ahead of
Anthropic's own top-tier Fable 5.1 at 85.3\%; Gemini 3.8 Flash sits around
82\%. Since Opus 5 sits below Fable in Anthropic's lineup, its position in our
results is at least consistent with that ordering --- though we note Anthropic
stopped publishing MRCR, RULER and needle-in-a-haystack figures from the Claude Opus
4.8 system card onwards, so Opus 5 has \emph{no} published deep-comprehension
long-context score and this inference rests on a proxy. Fiction.liveBench and
RULER have no published entries for any of our six models.

\paragraph{The compounding result has precedent.} \citet{laban2025multiturn},
decomposing over 200{,}000 simulated multi-turn conversations, report an
average 39\% performance drop across six generation tasks and attribute it
``primarily to increased unreliability ($+112\%$) rather than a loss of
aptitude ($-15\%$)''. That is the same decomposition our $p^{\log_2 N}$ fit
makes: the per-step task does not get harder, there are simply more steps to
fail at. \citet{kwa2025horizons} formalise a related structure in task
\emph{length}, finding 80\% task-completion horizons roughly $5\times$ shorter
than 50\% horizons --- the signature of compounding per-step reliability.
\citet{backlund2025} similarly find long-horizon agent failures uncorrelated
with context-window occupancy, pointing at reliability rather than memory.

\paragraph{Overthinking is documented.} Our finding that reasoning expenditure
varies $4.5\times$ at $N{=}1024$ with little relation to success, and that
Gemini at MEDIUM
outperformed models at HIGH, is consistent with \citet{gema2025inverse}, who
construct tasks where extending reasoning length \emph{degrades} accuracy.

The same pattern appears on grounded generation. Vectara's revised hallucination
leaderboard finds that reasoning models summarise short documents \emph{less}
faithfully than their non-reasoning predecessors: GPT-5, Claude Sonnet 4.5,
Grok-4 and Gemini-3-Pro all exceed 10\%, against 1.5\% for GPT-4o and 4.6\% for
Claude 3.5 Sonnet, with one Grok variant at 20.2\%
\citep{vectara2026,bang2025hallulens}. The proposed explanation is that
reasoning models invest effort in thinking through an answer that faithful
transcription does not require. That is a different task and a different
failure, but it is the same direction on the same axis, observed independently.

\paragraph{Lexical bisection is harder than it looks.} \citet{edman2024cute}
find that models ``seem to know the spelling of their tokens, yet fail to use
this information effectively'', because tokenisation obscures characters. That
only two of our six models use first-letter partitions --- and that those two
are the only two reaching $H(p)=1.000$ --- is therefore more notable than it
would be if such partitions were easy: it suggests unusually reliable
sub-token access rather than a strategy any model could adopt at will.

\paragraph{Asymmetric-information games and theory of mind.} Codenames has been
used as a benchmark with the same clue-giver/guesser asymmetry
\citep{stephenson2024codenames}, and reports the same difficulty we observe
when agents do not co-adapt. On theory-of-mind benchmarks, frontier models
trail humans substantially \citep{xu2024tombench}, and interactive
coordination tasks separate far more sharply than static belief probes ---
consistent with our finding that failures concentrate upstream of the final
inference (\S\ref{sec:errors}).

\paragraph{Content filtering on China-hosted APIs is well documented.}
\citet{pan2026censorship} measure refusal rates across 145 political prompts:
DeepSeek $\approx$36\%, Ernie 32\%, ChatGLM 10\%, against 0\% for GPT.
\citet{yang2025moderation} report 47\% refusal by DeepSeek across 1{,}360
queries. Independent testing has found that identical open weights score
79.8\% through a hosted API against 95.2\% run locally, with 76\% of the API
failures being blank responses --- direct evidence that the filtering is an
API-layer property rather than a property of the weights. Our observations
(\S\ref{sec:filter}) are consistent with this in kind: the subjects that were
refused --- Taiwan's navy, a purged intellectual, a detained Hong Kong activist
--- match the documented sensitivity pattern, and because Kimi K3 and GLM-5.3
are open-weight models, the same weights served elsewhere would likely not
refuse.

\paragraph{Cost decoupling.} Third-party per-task cost measurements show the
same loose coupling we find, with open-weight models achieving comparable index
scores at a fraction of the per-task cost of the most expensive frontier
offerings.

\medskip
\noindent In summary, the mechanisms this report invokes are independently documented,
which raises our confidence that the effects are real. The specific per-model
ordering rests on 68 games per model and should be replicated before it is
treated as a stable ranking.

\section{Limitations}
\label{sec:limits}

\paragraph{One document set per size.} Each size uses a single document set with
8 targets, confounding $N$ with that set's difficulty. This likely explains
non-monotonicity such as Claude Opus 5 at 75\% for $N{=}32$ and 12.5\% for $N{=}16$; with
$n=8$, Wilson intervals overlap almost everywhere. The cross-model comparison is
unaffected, being paired. Claims about the shape of the $N$-curve are weak;
claims about between-model differences are not.

\paragraph{Substituted documents.} Two arms did not run on the identical frozen
set. GLM-5.3's $N{=}1024$ set replaces one document and Kimi K3's replaces
three, in both cases because a provider content filter refused to process them
(\S\ref{sec:filter}). No substituted document was a target, $N$ and the round
budget are preserved, and each substitution is recorded in the affected result
files. The perturbation is $0.1\%$ and $0.3\%$ of the respective sets.

\paragraph{Non-deterministic arms.} Three of six providers do not permit
temperature control.

\paragraph{Prompt iteration, and which model the pilot used.} Prompts were
revised four times against observed mechanisms during a pilot, then frozen
before any reported arm was run. Several pilot instances involved the same
document, so overfitting to them is possible.

Every pilot game was played on Claude Opus 5, and all four revisions were
driven by observations from its games. Two of the four corrected defects in our
own instructions rather than model behaviour: the first answerer prompt forbade
world knowledge, and the first questioner prompt permitted questions that
presupposed the candidate set. The revisions were: the world-knowledge licence
came from its answer to the
Vocento/Europe question, the contrastive prohibition and its generalisation
from two of its questions, and the restoration of answerer reasoning from its
21\% Yes rate. The frozen prompts are therefore tuned to compensate for one
model's behaviour, and that model then finished last.

The direction of this confound runs against the headline result rather than
producing it. Tuning against Opus 5 should, if anything, have helped Opus 5.
\S\ref{sec:errors} adds to this. The world-knowledge licence exists
specifically to override a disposition toward document-only literalism that
Anthropic's disclosed training encourages, and Opus 5 retains the highest answer
error rate under the corrected wording. What we cannot rule out is the weaker
version, that prompts shaped around one model's idiosyncrasies fit the other
five better or worse than a neutrally derived set would. Deriving prompts from
a pilot on a model outside the evaluated set would remove this.

\paragraph{Context load is not equalised across models.} Our $N{=}1024$
prompt is $\approx$126{,}000 tokens, but that is a different fraction of each
model's trained context. Kimi K3 supports 1M tokens \citep{kimi2026k3}, so the
prompt is $\approx$13\% of its window; GLM-5's report describes a context
extended progressively to 200K during a dedicated mid-training phase
\citep{zhipu2026glm5}, making the same prompt $\approx$63\% of a window reached
by extension rather than trained natively. RULER \citep{hsieh2024} and NoLiMa
\citep{modarressi2025nolima} both find advertised context substantially exceeds
effective context. GLM-5.3 has both the steepest decay slope ($r=-0.63$) and
the worst $N{=}1024$ result (2/8) among the leading five, and extended-context
degradation is a credible alternative to any explanation we offer. Because its
weights are public, this is directly testable by serving it locally at reduced
context.

\paragraph{Architectural claims are unavailable for four of six models.} Only
Kimi K3 and GLM-5 publish technical reports. Anthropic, OpenAI, Google and xAI
disclose neither parameter counts, attention mechanisms, tokenizers, data
mixtures, nor post-training recipes for the models tested. Any mechanistic
account we offer for those four rests on published \emph{behavioural} posture
(\S\ref{sec:errors}) rather than on architecture, and should be read as
hypothesis rather than explanation. This is a particular limitation for the
model that performs worst here, which comes from the lab disclosing least.

\paragraph{One flagship per provider is an imperfect protocol.} Vendors' tier
structures are not commensurable. Claude Opus 5 is Anthropic's mid-priced
flagship with higher tiers above it, while Gemini 3.8 Flash is the newest
model in a line whose ``Pro'' tier has fallen behind it. We chose per provider
the model we judged most representative and most likely to be used; a different
defensible choice would change the ordering (\S\ref{sec:external}).

\paragraph{Effort levels are labels, not a scale.} The $4.5$--$11\times$ spread in
token expenditure at nominally identical settings means ``all models at high''
is a protocol, not a controlled variable. Gemini's documented exception at
MEDIUM compounds this.

\paragraph{Answer balance is an indirect estimate.} \S\ref{sec:bits} infers
information per question from response frequencies, which conflates question
balance with answerer bias. A direct measurement is proposed below.

\section{Future work}
\label{sec:future}

The harness is released. The extensions below are the ones we would
prioritise.

\paragraph{Open-weight ablations.} Two of the six models have public weights,
which makes several of this report's open questions decidable at low cost by
serving them locally.
(i) \emph{Separating API filtering from model behaviour}: re-submit the three
refused paragraphs (\S\ref{sec:filter}) to locally served GLM-5.3 and Kimi K3.
If they process cleanly, refusal is confirmed as a serving-stack property, as
independent testing of other Chinese-hosted models suggests.
(ii) \emph{Separating context load from attention mechanism}: run GLM-5.3 at
$N \le 256$, far below its context limit. If its decay slope flattens to match
Kimi K3's, context load rather than attention type is the driver.
(iii) \emph{Testing the aggregation hypothesis directly}: evaluate both models
on RULER's aggregation subtasks at matched context length. If the selective
sparse architecture underperforms the linear-hybrid one specifically on
aggregation but not retrieval, \S\ref{sec:longctx}'s sharpened claim is
supported.

\paragraph{Locating the ``No'' bias.} The answerer task can be run standalone,
without the game: present one document and one question whose answer is stated
in the text, and vary the question's polarity so that half the items have
ground-truth Yes and half No. Per-model error rates by polarity would establish
whether the bias reported in \S\ref{sec:errors} is specific to this task
framing, to the one-word output constraint, or to the model. It would also
scale to far more items than 34, which is all our logs contain for the affected
model.

\paragraph{Effort measured rather than declared.} Vendor effort labels are not
comparable (\S\ref{sec:limits}). Plotting accuracy against \emph{emitted}
reasoning tokens rather than the nominal setting would test whether our
``Gemini at MEDIUM beats models at HIGH'' comparison was ever like-for-like.

\paragraph{Other modalities.} Nothing in the game requires the candidates to be
text. Replacing each document with an image, a chart, an audio clip, or a short
video would preserve the structure exactly --- the questioner surveys $N$ items
and partitions them, the answerer sees one and adjudicates --- while changing
what the partition must be built from. This makes it a natural multimodal
reasoning benchmark, and one with a property such benchmarks usually lack: the
questioner must form a \emph{global} judgement over the whole candidate set,
not merely describe individual items. A mixed-modality variant, where the
questioner sees images and the answerer receives captions or vice versa, would add a
representational gap to the informational one, and would test whether the
contrastive failures of \S\ref{sec:comm} become more common when the two agents
do not share a modality.

\paragraph{Adjudicating the full partition.} \S\ref{sec:errors} adjudicates two
documents per game, which suffices to separate answer errors from
discrimination failures but not to measure \emph{split quality} --- how evenly
each question divides the viable set. That needs the full partition, which is
$O(N)$ per round and therefore roughly 400$\times$ more adjudications at
$N{=}1024$. It would settle whether the round-position effect of
\S\ref{sec:aggregation} is real or an artefact of assuming even splits.

\paragraph{Measuring agreement with intent rather than with truth.} Our
adjudication asks whether an answer is correct. A communication account needs
whether the answerer's judgement matches what the questioner meant. The
questioner's reasoning traces often state the intended partition, so the two can
be compared from existing logs, distinguishing an answerer that is wrong from
one that answered a different question than the one asked.

\paragraph{Context-asymmetry ablation.} Ask the same model the same question
about the same document twice: once as answerer (document only) and once with
the full set visible. Divergence isolates the cost of the asymmetry itself,
which is precisely the multi-agent quantity.

\paragraph{Cross-model pairing.} Running questioner and answerer on
\emph{different} providers would separate a model's ability to ask
well-specified questions from its ability to adjudicate them, and would test
whether communication protocols transfer across model families.

\paragraph{Attention-architecture correlation.} If the round-1 aggregation
deficit reflects how a model attends over long unconstrained inputs, the metric
should correlate with disclosed architectural choices --- sparse,
sliding-window, or block-routed attention --- across a wider model set than we
test here. Open-weight models would permit the direct version of this test.

\paragraph{Isolating the survey.} Round 1 still confounds aggregation with
question generation. A cleaner probe would ask a model directly to estimate what
fraction of a document set satisfies a stated property, scoring against ground
truth. That would establish whether the deficit is in surveying the context or
in choosing what to ask about it.

\paragraph{Multiple document sets per size.} The principal fix for
\S\ref{sec:limits}: eight games per cell on a single document set confounds
set difficulty with $N$.

\paragraph{Longer or shorter documents.} Our paragraphs average 123 tokens.
Single sentences would make partitions coarser and harder to balance; full
articles would test aggregation over a far larger context at the same $N$.
Both are one line of the corpus builder.

\section{Cost}
\label{sec:cost}

We report spend in full, since the practicality of this kind of evaluation for
an unfunded researcher is part of what the report is meant to establish.

\begin{table}[h]
\centering\small
\begin{tabular}{lrrrrr}
\toprule
Arm & Games & Cost & \$/game & \$/win & Wall clock \\
\midrule
GLM-5.3            & 68 & \$16.26  & 0.24 & \textbf{0.36} & 7.2\,h \\
Gemini 3.8 Flash   & 68 & \$21.35  & 0.31 & \textbf{0.39} & 3.6\,h \\
Claude Opus 5      & 68 & \$66.21  & 0.97 & 2.36 & 12.7\,h \\
Kimi K3            & 68 & \$68.72  & 1.01 & 1.23 & 20.3\,h \\
Grok 4.6           & 68 & \$70.02  & 1.03 & 1.37 & 25.3\,h \\
GPT-5.6 Sol        & 68 & \$120.60 & 1.77 & 2.46 & 3.9\,h \\
\midrule
\textbf{Total}     & \textbf{408} & \textbf{\$363.15} & & & \\
\midrule
Discarded: Gemini at HIGH (\S\ref{sec:stopreason}) & 68 & $\approx$\$47 & & --- & \\
Discarded: DeepSeek V4-Pro pilot & 2 & $\approx$\$1 & & --- & \\
Pilots, probes and debugging & --- & $\approx$\$20 & & --- & \\
\bottomrule
\end{tabular}
\end{table}

\begin{figure}[tbp]
\centering
\begin{tikzpicture}
\begin{axis}[
  width=11cm, height=6.5cm,
  xlabel={arm cost (USD)}, ylabel={win rate},
  xmin=0, xmax=135, ymin=0.30, ymax=0.92,
  grid=major, nodes near coords, point meta=explicit symbolic,
  every node near coord/.append style={font=\scriptsize, anchor=south west},
]
\addplot[only marks, mark=*, mark size=2.5pt, blue] coordinates {
  (16.26,0.66) [GLM-5.3]
  (21.35,0.81) [Gemini 3.8]
  (66.21,0.41) [Opus 5]
  (68.72,0.82) [Kimi K3]
  (70.02,0.75) [Grok 4.6]
  (120.60,0.72) [GPT-5.6]
};
\end{axis}
\end{tikzpicture}
\caption{Arm cost against win rate. The two are uncorrelated ($r=-0.05$): the two cheapest arms are among the three most accurate.}
\label{fig:cost}
\end{figure}
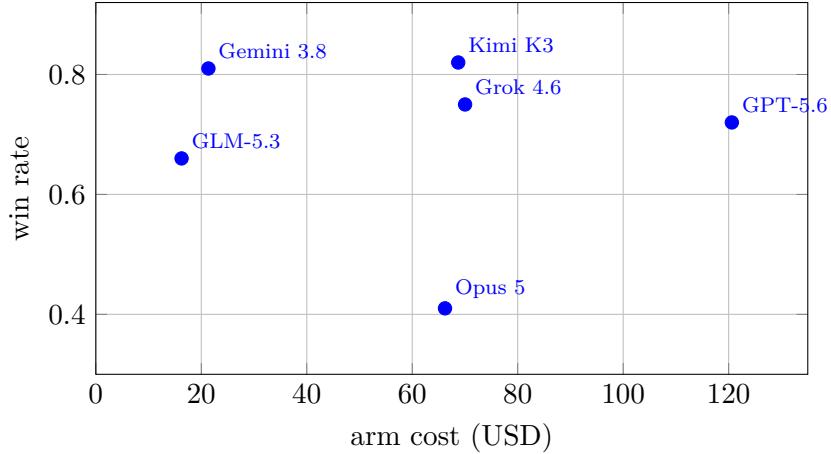

Four observations. First, cost and performance are unrelated ($r=-0.05$):
the two cheapest arms are also two of the three best, and GPT costs $7\times$
GLM for a lower win rate. Second, cost and wall clock are unrelated to each
other --- GPT is the most expensive arm and among the fastest, while Grok and
Kimi are mid-priced and took 20--25 hours. Wall clock, not money, was the
binding constraint on this project. Third, the discarded Gemini arm is listed
deliberately: an engineering error that survives to a completed arm costs real
money, which is the practical argument for the diagnostics of \S\ref{sec:eng}.
Fourth, the whole study cost less than a single day of a researcher's time at
commercial rates, which we note because the barrier to work of this kind is
often assumed to be compute.

\section{Reproducibility}

The corpus, document-set manifest, prompts, and all per-game logs are released.
Every result file records the prompt version hash, schema version, corpus hash,
the model identifier echoed by the API, per-call token usage including cached
and reasoning tokens, raw response text, reasoning traces, and the reasoning
configuration actually used. The runner refuses to treat a game as complete if
its prompt or schema version differs from the current one.

Code, corpus and all per-game logs:
\url{https://github.com/ppotash/logn-questions}

\end{document}